\documentclass{article} 
\usepackage{iclr2015,times}
\usepackage{hyperref}
\usepackage{url}

\usepackage{amsfonts,amstext,amsmath,amssymb}

\usepackage{booktabs}
\newcommand{\ra}[1]{\renewcommand{\arraystretch}{#1}}
\usepackage{array}

\usepackage{graphicx}
\usepackage{caption}
\usepackage{subcaption}

\title{N-gram-Based Low-Dimensional Representation for Document Classification}

\author{
R\'emi Lebret\\
Idiap Research Institute, Martigny, Switzerland \\
Ecole Polytechnique F\'ed\'erale de Lausanne (EPFL), Lausanne, Switzerland\\
\texttt{remi@lebret.ch} \\
\And
Ronan Collobert\thanks{All research was conducted at the Idiap Research Institute, before Ronan Collobert joined Facebook AI Research.}\\
Facebook AI Research, Menlo Park, CA, USA\\
Idiap Research Institute, Martigny, Switzerland \\
\texttt{ronan@collobert.com} \\
}

\iclrfinalcopy 

\begin{document}

\maketitle

\begin{abstract}
The bag-of-words (BOW) model is the common approach for classifying documents, where words are used as feature for training a classifier.
This generally involves a huge number of features.
Some techniques, such as Latent Semantic Analysis (LSA) or Latent Dirichlet Allocation (LDA), have been designed to summarize documents in a lower dimension with the least semantic information loss.
Some semantic information is nevertheless always lost, since only words are considered. 
Instead, we aim at using information coming from $n$-grams to overcome this limitation, while remaining in a low-dimension space.
Many approaches, such as the Skip-gram model, provide good word vector representations very quickly.
We propose to average these representations to obtain representations of $n$-grams.
All $n$-grams are thus embedded in a same semantic space. 
A $K$-means clustering can then group them into semantic concepts.
The number of features is therefore dramatically reduced and documents are then represented as bag of semantic concepts. 
We show that this model outperforms LSA and LDA on a sentiment classification task, and yields similar results than a traditional BOW-model with far less features.
\end{abstract}

\section{Introduction}

Text document classification aims at assigning a text document to one or more classes.
Successful methods are traditionally based on \emph{bag-of-words} (BOW). 
Finding discriminative keywords is, in general, good enough for text classification.
Given a dictionary of words $\mathcal{D}$ to consider, documents are represented by a $|\mathcal{D}|$-dimensional vector (the bag of its words). 
Each dimension is either a binary value (present or not in the document) or a word occurrence frequency.
Some term weightings ({\it e.g.} the popular td-idf) have also been defined to reflect how discriminative a word is for a document.
These are considered as features for training a classifier. 
Naive Bayes (NB) and Support Vector Machine (SVM) models are often the first choices.
One limitation of the bag-of-words model is that the discriminative words are usually not the most frequent ones.
A large dictionary of words needs to be defined to obtain a robust model. 
Classifiers then have to deal with a huge number of features, and thus become time-consuming and memory-hungry.

Some techniques have been proposed to reduce the dimensionality and represent documents in a low-dimensional semantic space.
Latent Semantic Analysis (LSA) \citep{Deerwester90} uses the term-document matrix and a singular value decomposition (SVD) to represent terms and documents in a new low-dimensional space. 
Latent Dirichlet Allocation (LDA) \citep{Blei2003} is a generative probabilistic model of a corpus. Each document is represented as a mixture of latent topics, where each topic is characterized by a distribution over words. By defining $K$ topics, documents can then be represented as $K$-dimensional vectors.
\citet{Pessiot09} also proposed probabilistic models for unsupervised dimensionality reduction in the context of document clustering. 
They make the hypothesis that words occuring with the same frequencies in the same document are semantically related. Based on this assumption, words are partioned into word topics.
Document are then represented by a vector where each feature corresponds to a word-topic representing the number of occurrences of words from that word-topic in the document.
Other techniques have tried to improve text document clustering by taking into account relationships between important terms.
Some have enriched document representations by integrating core ontologies as background knowledge \citep{StaabH03}, or with Wikipedia concepts and category information \citep{Hu2009}. Part-of-speech tags have also been used to disambiguate words \citep{Sedding2004}.

All these techniques are based on words alone, which raises another limitation. 
A collection of words cannot capture phrases or multi-word expressions, while $n$-grams have shown to be helpful features in several natural language processing tasks \citep{Tan2002,Lin2009,Wang12}.
$N$-gram features are not commonly used in text classification, probably because the dictionary $\mathcal{D}^n$ tends to grow exponentially with $n$. 
Phrase structure extraction can be used to identify only $n$-grams which are phrase patterns, and thus limit the dictionary size.
However, this adds another step to the model, making it more complex.
To overcome these barriers, we propose that documents be represented as a \emph{bag of semantic concepts}, where $n$-grams are considered instead of only words.
Good word vector representations are obtained very quickly with many different recent approaches \citep{Mikolov2013,Mnih2013,Lebret14,pennington2014glove}.
\citet{MikolovICLR2013} also showed that simple vector addition can often produce meaningful results, such as \emph{king - man + woman $\approx$ queen}.
By leveraging the ability of these word vector representations to compose, representations for $n$-grams are easily computed with an element-wise addition.
Using a clustering algorithm such as $K$-means, those representations are grouped into $K$ clusters which can be viewed as \emph{semantic concepts}.
Text documents are now represented as bag of semantic concepts, with each feature corresponding to the presence or not of $n$-grams from the resulting clusters.
Therefore, more information is captured while remaining in a low-dimensional space.
As Mikolov et al's Skip-gram model and $K$-means are highly parallelizable, this model is much faster to compute than LSA or LDA. 
The same classifiers as with BOW-based models are then applied on these bag of semantic concepts.
We show that such model is a good alternative to LSA or LDA to represent documents and yields even better results on movie review tasks.

\section{A bag of semantic concepts model}
\label{model}
The model is divided into three steps: (1) vector representations of $n$-grams are obtained by averaging pre-trained representations of its individual words; (2) $n$-grams are gouped into $K$ semantic concepts by performing $K$-means clustering on all $n$-gram representations; (3) documents are represented by a bag of $K$ semantic concepts, where each entry depends on the presence of $n$-grams from the concepts defined in the previous step.

\subsection{$N$-gram representation}
The first step of the model is to generate continuous vector representations $\mathrm{x}_{w}$ for each word $w$ within the dictionary $\mathcal{D}$.
Leveraging recent models, such as the Skip-gram \citep{Mikolov2013} or GloVe \citep{pennington2014glove} models, they are trained over a large corpus of unlabeled data in an efficient manner. These models are indeed highly parallelizable, which helps to obtain these representations very quickly.
Word representations are then summed to generate $n$-gram representations:
\begin{equation}
\frac{1}{n} \sum_{i=1}^n \mathrm{x}_{w_i}\,.
\end{equation}
These representations are vectors which keep the semantic information of $n$-grams with different $n$ in the same dimensionality.
Distances between them are thus computable. 
It allows the use of a $K$-means clustering for grouping all $n$-grams into $K$ classes.

\subsection{$K$-means clustering}
$K$-means is an unsupervised learning algorithm commonly used to automatically partition a data set into $K$ clusters.
Considering a set of $n$-gram representations $\mathrm{x}_i \in \mathbb{R}^m$, the algorithm will determine a set of $K$ centroids $\gamma_k \in \mathbb{R}^m$, so as to minimize the average distance from each representation to its nearest centroid:
\begin{equation}
\sum_{i} \, || \mathrm{x}_i - \gamma_{\sigma_i}||^2\,,\,\,\,\text{where } \sigma_i = \operatorname*{argmin}_{k}\, ||\mathrm{x}_i - \gamma_k||^2 \,.
\end{equation}
The limitation due to the size of the dictionary is therefore overcomed. 
By setting $K$ to a low value, documents can also be represented by more compact vectors than with a bag-of-words model, while keeping all the meaningful information.

\subsection{Document representation}
\label{docrep}
Denoting $D=(d_1,d_2,\ldots,d_L)$ a set of text documents, where each document $d_i$ contains a set of $n$-grams.
First, each $n$-gram is embedded into a common vector space by averaging its word vector representations.
The resulting $n$-grams representations are assigned to clusters using the centroids $\gamma_k$ defined by the $K$-means clustering.
Documents $d_i$ are then represented by a vector of $K$ features, $\mathrm{f}_i \in \mathbb{R}^K$.
Each entry $\mathrm{f}_i^k$ usually corresponds to the frequency of $n$-grams from the $k^{\text{th}}$ cluster within the document $d_i$.
The set of text documents is then defined as $\hat{D} = \{(\mathrm{f}_i,y_i) |\, \mathrm{f}_i \in \mathbb{R}^K, y_i \in \{-1,1\}\}_{i=1}^L$.

\paragraph{With NB features.}
For certain type of document, such as movie reviews, the use of Naive Bayes features can improve the general performance \citep{Wang12}.
Success in sentiment analysis relies mostly on the capability of the models to detect negative and positive $n$-grams in a document.
A proper normalization is then calculated to determine how important each $n$-gram is for a given class $y$.
We denote $\mathrm{ngm}=(ngm_1,\ldots,ngm_N)$ a set of count vectors for all $n$-grams contained in $D$, $ngm_i \in \mathbb{R}^{L}$.
$ngm_t^i$ represents the number of occurence of the $n$-gram $t$ in the training document $d_i$.
Defining count vectors as $\mathrm{p} = 1 + \sum_{i:y_{i}=1} \mathrm{ngm}^i$ and $\mathrm{q} = 1 + \sum_{i:y_{i}=-1} \mathrm{ngm}^i$, 
a log-count ratio is calculated to determine how important $n$-grams are for the classes $y$:
\begin{equation}
\mathrm{r} = \operatorname*{log} \left( \frac{\mathrm{p}/||\mathrm{p}||_1}{\mathrm{q}/||\mathrm{q}||_1} \right)\,,\, \text{with } \mathrm{r} \in \mathbb{R}^N\,.
\end{equation}

Because $n$-grams are in clusters, we extract the maximum absolute log-count ratio for every cluster: 
\begin{equation}
\mathrm{\tilde{f}}_i^k = \operatorname*{argmax}_{r_t}\, |r_t|\,, \,\forall t \in k, \, ngm_t^i>0
\end{equation}

These document representations can then be used for several NLP tasks such as classification or information retrieval.
As for BOW-based models, this model is particulary suitable for linear SVM.

\section{Experiments with sentiment analysis}

Sentiments can have a completely different meaning if $n$-grams are considered instead of words.
A classifier might leverage a bigram such as ``not good'' to classify a document as negative, while this would probably fail if only unigrams (words) were considered.
We thus benchmark the bag of semantic concepts model on sentiment analysis.

\subsection{IMDB movie reviews datasets}

Datasets from IMDB have the nice property of containing long documents.
It is thus valuable to considerer $n$-grams in such a framework.
We did experiments with small and large collections of reviews. 
We can thus analyse how well our model compares against classical models, for different dataset sizes.

\subsubsection{\citet{PangLee04}}
The collection consists of 1,000 positive and 1,000 negative processed reviews\footnote{Available at \small{\url{http://www.cs.cornell.edu/people/pabo/movie-review-data/}}.}. So a random guess yields 50\% accuracy.
The authors selected only reviews where rating was expressed either with stars or some numerical value.
To avoid domination of the corpus by a small number of prolific reviewers, they imposed a limit of fewer than 20 reviews per author per sentiment category.
As there is no test set, we used 10-fold cross-validation.

\subsubsection{\citet{MaasDPHNP11}}
The collection consists of 100,000 reviews\footnote{Available at \small{\url{http://www.andrew-maas.net/data/sentiment}}.}.
It has been divided into three datasets: training and test sets (25,000 labeled reviews each), and 50,000 unlabeled training reviews. 
It allows no more than 30 reviews per movie. It contains an even number of positive and negative reviews, so randomly guessing yields 50\% accuracy. Only highly polarized reviews have been considered. A negative review has a score $\leq$ 4 out of 10, and a positive review has a score $\geq$ 7 out of 10.

\subsection{Experimental setup}

We first learn word vector representations over a large corpus of unlabeled text. This step could however be skipped by taking existing pre-trained word representations\footnote{Different pre-trained word vector representations are available at \small{\url{https://code.google.com/p/word2vec/}}, \small{\url{http://stanford.edu/~jpennin/}} or \small{\url{http://lebret.ch/words/}}.} instead of learning them from scratch.
By following the three steps described in Section \ref{model}, movie reviews are then represented as bags of semantic concepts.
These representations are finally used for training a linear SVM to classify sentiment.

\subsubsection{Learning word representation over large corpora}
\label{wordrep}
Our English corpus is composed of the entire English Wikipedia\footnote{We took the January 2014 version.}, the Reuters corpus and the Wall Street Journal (WSJ) corpus. We consider lower case words and replace digits with a special token. The resulting text is tokenized using the Stanford tokenizer. The final data set contains about 2 billion words.
Our dictionary $\mathcal{D}$ consists of all the words appearing at least one hundred times. 
This results in a 202,255 words dictionary.
We then train a Skip-gram model to get word representation in a 100-dimensional vector. This dimension is intentionally quite low to speed up the clustering afterwards.
As other hyperparameters, we use a fixed learning rate of 0.01, a context size of 5 phrases, Negative Sampling with 5 negative samples for each positive sample, and a subsampling approach with a threshold of $10^{-5}$

\subsubsection{Bag of semantic concepts for movie reviews}

\paragraph{Computing $n$-gram representations.}
We consider $n$-grams up to $n=3$.
Only $n$-grams with words from our dictionary are considered for both datasets.\footnote{Our English corpus is not large enough to cover all the words present in the IMDB datasets. We thus use the same 1-gram dictionary with the other methods.}
This results in a set of 34,360 1-gram representations, 419,918 2-gram representations, and 921,837 3-gram representations for the Pang and Lee's dataset. 
And 67,847 1-gram representations, 1,842,461 2-gram representations, and 5,724,871 3-gram representations for the Maas et al.'s dataset.
Because $n$-gram representations are computed by averaging representations of its word, all $n$-grams are also represented in a 100-dimensional vector.

\paragraph{Partitioning $n$-grams into semantic concepts.}
Because $n$-grams are represented in a common vector space, similarities between $n$-grams of different length can be computed. 
To evaluate the benefit of adding $n$-grams for sentiment analysis, we define semantic concepts with different combinations of $n$-grams: (1) only 1-grams (i.e. clusters of words), (2) only 2-grams, (3) only 3-grams, (4) with 1-grams and 2-grams, and (5) with 1-grams, 2-grams and 3-grams.
Each of these five sets of $n$-gram representations are then partitioned in $K=\{100,200,300\}$ clusters with the $K$-means clustering. The centroids $\gamma_k \in \mathbb{R}^{100}$ are obtained after 10 iterations of the algorithm.

\paragraph{Movie review representations.}
Movie reviews are then represented as bags of semantic concepts with naive bayes features as described in Section \ref{docrep}. The log-count ratio for each $n$-gram is calculated on the training set for both datasets.

\subsubsection{Comparison with other methods}

We compare our models with two classical techniques for representing text documents in a low-dimensional vector space: LSA and LDA. 
Both methods use the same 1-gram dictionaries than with the bag of semantic concepts model with $K=\{100,200,300\}$.
In the framework of Maas et al.'s dataset, LSA and LDA benefit from the large set of unlabeled reviews.  

\paragraph{Latent Sentiment Analysis (LSA) \citep{Deerwester90}.} 
Let $X \in \mathbb{R}^{|\mathcal{D}| \times L}$ be a matrix where each element $X_{i,j}$ describes the log count ratio of words $i$ in document $j$, with $L$ the number of training documents and $\mathcal{D}$ the dictionary of words (i.e. 34,360 for Pang and Lee's dataset, 67,847 for Maas et al's dataset). 
By applying truncated SVD to the log-count ratio matrix $X$, we thus obtain semantic representations in a $K$-dimensional space for movie reviews.

\paragraph{Latent Dirichlet Allocation (LDA) \citep{Blei2003}.} 
We train the $K$-topics LDA model using the code released by \citet{Blei2003}\footnote{Available at \small{\url{http://www.cs.princeton.edu/~blei/lda-c/}}.}. We leave the LDA hyperparameters at their default values.
Like our model, LDA extracts $K$ topics (i.e. semantic concepts) and assigns words to these topics. 
Considering only the words in documents, we thus apply the method described in Section \ref{docrep} to get document representations.
A movie review $d_i$ is then represented in a $K$-dimensional vector, where each feature $\tilde{f}^k_i$ is the maximum absolute log-count ratio for the $k^{th}$ topic.

\subsubsection{Classification using SVM}
Having representations of movie reviews in a $K$-dimensional vector, a classifier is trained to determine whether a given review is positive or negative.  
Given the set of training documents $\tilde{D} = \{(\mathrm{\tilde{f}}_i,y_i) |\, \mathrm{\tilde{f}}_i \in \mathbb{R}^K, y_i \in \{-1,1\}\}_{i=1}^L$, we picked a linear SVM as a classifier, trained using the LIBLINEAR library \citep{Fan2008}:

\begin{equation}\label{eq:svm}
\operatorname*{min}_{\mathrm{w}}\,\,\,\, \frac{1}{2} \mathrm{w}^T\mathrm{w} + C\sum_i \operatorname*{max}(0, 1 - y_i \mathrm{w}^T \mathrm{\tilde{f}}_i)^2\,,
\end{equation}
with $\mathrm{w}$ the weight vector, and $C$ a penalty parameter. 

\subsection{Results}

\begin{table}
\begin{center}
\begin{tabular}{@{}rr>{\centering\arraybackslash}p{1.2cm}>{\centering\arraybackslash}p{1.2cm}>{\centering\arraybackslash}p{1.2cm}r>{\centering\arraybackslash}p{1.2cm}>{\centering\arraybackslash}p{1.2cm}>{\centering\arraybackslash}p{1.2cm}@{}}\hline\toprule
 & & \multicolumn{3}{c}{\bf Pang and Lee, 2004} & \phantom{abcdef} & \multicolumn{3}{c}{\bf Maas et al., 2011} \\
 \cmidrule{3-5}  \cmidrule{7-9} 
{$K=$} &  \phantom{abcdef} & \multicolumn{1}{c}{100} & \multicolumn{1}{c}{200} & \multicolumn{1}{c}{300} & \phantom{abcdef} & \multicolumn{1}{c}{100} & \multicolumn{1}{c}{200} & \multicolumn{1}{c}{300} \\\midrule
\\ 
LDA & & 76.20 & 77.10 & 76.80 & & 85.43 & 85.45 & 84.40\\
LSA & & 81.60 & 82.55 & 83.75 & & 85.82 & 86.63 & 86.88 \\\midrule
1-grams & & 81.60 & 82.60 & 82.70 & & 84.51 & 84.76 & 85.54 \\
2-grams & & 82.30 & 82.25 & 83.15 & & 88.02 & 88.06 & 87.87 \\
3-grams & & 73.85 & 73.05 & 72.65 & & 87.41 & 87.46 & 87.22 \\
1+2-grams & & 83.85 & {\bf 84.00} & {\bf 84.00} & & 88.10 & 88.19 & 88.18 \\
1+2+3-grams & & 82.45 & 83.05 & 83.05 & & 88.39 & 88.46 & {\bf 88.55} \\\bottomrule
\hline
\end{tabular}
\caption{Classification accuracy on both movie review tasks with $K=\{100,200,300\}$ number of features.}
\label{tab:bon}
\end{center}
\end{table}

The overall results summarized in Table \ref{tab:bon} show that the bag of semantic concepts approach outperforms the traditionnal LDA and LSA approaches to represent documents in a low-dimensional space.
Good performance is achieved even with only 100 clusters, where LSA needs more clusters to improve.
We also denote that our approach performs well on a small dataset, where LDA fails.
A significant increase is observed when using 2-grams instead of 1-grams. However, using only 3-grams hurts the performance. 
The best results are obtained using a combination of $n$-grams, which confirms the benefit of the method.
That also means that word vector representations can be combined while keeping relevant semantic information. 
This is illustrated in Table \ref{tab:clus} where semantically close $n$-grams are in the same cluster.
We can see that the model is furthermore able to clearly separate antonyms, which is a good asset for sentiment classification.
The results are also very competitive with a traditional BOW-model. 
Using the same 1-gram dictionary and a linear SVM classifier with the naive bayes features, BOW-model achieves 83\% accuracy for Pang and Lee's dataset, and 88.58\% for Maas et al's dataset.
Our model therefore performs better with about 344 times less features for the first dataset, and yields similar result with about 678 times less features for the second one.

\subsection{Computation time}

The bag of semantic concepts model can leverage information coming from $n$-grams to improve sentiment classification of documents. 
This model has also the nice property to build document representations in an efficient and timely manner. 
The most time-consuming and costly process step in the model is the $K$-means clustering, especially when dealing with millions of $n$-gram representations. However, this step can be done very quickly with low memory by using mini-batch $K$-means method.
Computation times for generating 300-dimensional representations are reported in Table \ref{tab:time}.
All experiments have been run on single CPU core Intel i7 2600K 3.4 GHz. Despite the fact that single CPU has been used for this benchmark, the three steps of the model are highly parallelizable. The recorded times could thus be divided by the number of CPUs available.
We see that representations can be computed in less than one minute with only 1-gram dictionary. About 10 minutes are necessary when adding 2-grams, and about 40 minutes by adding 3-grams.
In comparison, LDA needs six hours for extracting 100 topics and three days for 300 topics.
Our model is also very competitive with LSA which takes 540 seconds to generate 300-dimensional document representations.
However, adding 2-grams and 3-grams to perform a LSA would be extremely time-consuming and memory-hungry while our model can handle it.

\begin{table}[h]
\begin{center}
\ra{1.3}
\begin{tabular}{rrrrrr}\hline\toprule
& \multicolumn{1}{r}{\bf 1-grams} & \multicolumn{1}{r}{\bf 2-grams}& \multicolumn{1}{r}{\bf 3-grams}& \multicolumn{1}{r}{\bf 1+2-grams}& \multicolumn{1}{r}{\bf 1+2+3-grams} \\\midrule
$N$-gram Representations & 0 & 43.00 & 164.34 & 43.00 & 207.34 \\
$K$-means &  14.18 & 291.62 & 747.90 & 302.34 & 1203.99 \\
Document Representations &  36.45 & 173.48 & 494.06 & 343.29 & 949.01 \\\midrule
Total & 50.63 & 508.10 & 1406.30 & 688.63 & 2360.34  \\\bottomrule
 \hline
\end{tabular}
\caption{Computation time for building movie review representations with $K=300$ semantic concepts. Time is reported in seconds.}
\label{tab:time}
\end{center}
\end{table}

\subsection{Inferring semantic concepts for unseen $n$-grams}

\begin{table}[h]
\begin{center}
\ra{1.3}
\begin{tabular}{@{}lrrclrr@{}}\hline\toprule
  \multicolumn{1}{c}{\bf good} & \phantom{ab}  & \multicolumn{1}{c}{\bf not good} & \phantom{abcde} & \multicolumn{1}{c}{\bf enjoy} & \phantom{ab}  & \multicolumn{1}{c}{\bf did n't enjoy} \\
  \cmidrule{1-3} \cmidrule{5-7} 
  \multicolumn{1}{c}{$k$=269} & & \multicolumn{1}{c}{$k$=297} & \phantom{abc} & \multicolumn{1}{c}{$k$=160} & & \multicolumn{1}{c}{$k$=108} \\\midrule
  \\
   nice one & & sufficiently bad &  & entertain & & sceptics\\
   liked here &  & not liked & & adored them & & did n't like \\
   is pretty nice &  & is far worse & & enjoying  & & n't enjoy any \\
   the greatest thing   &  & not that greatest & & watched and enjoy & & valueless\\\bottomrule
 \hline
\end{tabular}
\caption{Selected pairs of antonyms and their cluster number. Here, $n$-grams from Maas et al's dataset have been partitioned into 300 clusters. Each $n$-gram is accompagnied with a selection of others from its cluster.}
\label{tab:clus}
\end{center}
\end{table}

Another drawback of classical models is that they cannot deal with unseen words. Only words present in the training documents are used to infer representation for a new text document.
Unlike these models, our model can easily assign semantic concepts for new $n$-grams.
Because $n$-gram representations are based on its word vector representations, a new $n$-gram vector representation can be calculated if a representation is available for each of its words.
This new representations is then assigned to the nearest centroid $\gamma_k$, which determines its semantic concept.
With a small training set, this is a valuable asset when compared to other models.

\section{Conclusion}

Word vector representations can be quickly obtained with recent techniques such as the Skip-gram model.
$N$-grams with different length $n$ can then be embedded in a same dimensional vector space with a simple element-wise addition.
This makes it possible to compute distances between $n$-grams, which can have many applications in natural language processing.
We therefore proposed a bag of semantic concepts model to represent documents in a low-dimensional space.
These semantic concepts are obtained by performing a $K$-means clustering which partition all $n$-grams into $K$ clusters. 
This model has several advantages over classical approaches for representing documents in a low-dimensional space: it leverages semantic information coming from $n$-grams; it builds document representations with low resource consumption (time and memory); it can infer semantic concepts for unseen $n$-grams.
Furthermore, we have shown that such model is suitable for document classification.
Competitive performance has been reached on binary sentiment classification tasks, where this model outperforms traditional approaches. 
It also attained similar results to traditional bag-of-words with considerably less features.

\section*{Acknowledgments}

This work was supported by the HASLER foundation through the grant ``Information and Communication Technology for a Better World 2020'' (SmartWorld).

\bibliographystyle{iclr2015}
\bibliography{iclr2015}

\end{document}